\documentclass[11pt, twocolumn]{article}

\usepackage[margin=0.75in]{geometry}
\usepackage{amsmath, amsfonts, amsthm, amssymb}
\usepackage{mathtools, bm}
\usepackage{url}
\usepackage{color}
\usepackage{subfig}
\usepackage[round, authoryear]{natbib}
\usepackage[colorlinks=true, allcolors=blue]{hyperref}
\usepackage{booktabs}
\usepackage{array}
\usepackage{multicol}

\title{\rule{\linewidth}{3pt} \vspace{0.5mm} \\
\textbf{\huge{\textsf{CoviHawkes}: Temporal Point Process and Deep
    Learning based Covid-19 forecasting for India}} \\
\rule{\linewidth}{1pt}}

\author{\textbf{Ambedkar Dukkipati}, \textbf{Tony Gracious}, and \textbf{Shubham Gupta} \vspace{4mm} \\
Department of Computer Science and Automation, \\
Indian Institute of Science, Bangalore, INDIA. \vspace{1mm}\\
\texttt{[ambedkar, tonygracious, shubhamg]@iisc.ac.in}}

\date{}

\begin{document}

\maketitle


\begin{abstract}
    Lockdowns are one of the most effective measures for containing the spread of a pandemic. Unfortunately, they involve a heavy financial and emotional toll on the population that often outlasts the lockdown itself. This article argues in favor of ``local'' lockdowns, which are lockdowns focused on regions currently experiencing an outbreak. We propose a machine learning tool called \textsf{CoviHawkes} based on temporal point processes, called CoviHawkes that predicts the daily case counts for Covid-19 in India at the national, state, and district levels. Our short-term predictions ($<30$ days) may be helpful for policymakers in identifying regions where a local lockdown must be proactively imposed to arrest the spread of the virus. Our long-term predictions (up to a few months) simulate the progression of the pandemic under various lockdown conditions, thereby providing a noisy indicator for a potential third wave of cases in India. Extensive experimental results validate the performance of our tool at all levels.
\end{abstract}


\section{Introduction}
\label{section:introduction}
We live in strange times. The Covid-19 pandemic has disrupted the lives and livelihood of people in unfathomable, if not unprecedented, ways. It has brought the devastating effects of social isolation to the center stage and challenged the very fabric of our society and its economy. While public safety measures like lockdowns help arrest the spread of the virus, unless imposed early on, their primary purpose is to delay the inevitable to allow healthcare systems to prepare themselves for the looming crises. As such, these lockdowns should be treated as strategic decisions, and data-aware methodologies should guide their implementation. This article provides technical details of a deep learning-based tool called \textsf{CoviHawkes}\footnote{\url{https://sml.csa.iisc.ac.in/covihawkes}} that forecasts India's daily case counts at the national, state, and district levels. The predictions made by our tool may help the policymakers identify vulnerable regions to enact ``local'' lockdowns proactively.

Lockdowns have a disproportionate impact on different people. For example, while employees working in an IT firm can often continue their job remotely, taxi drivers and food delivery personnel lose their daily livelihood. The spending power of the consumers reduces as more people lose their income, which damages the overall economy and leads it into a vicious circle. The consequences of these lockdowns often outlast their duration by a wide margin. Combined with the fact that the infection outbreaks are almost always localized, it arguably makes more sense to enforce lockdowns in regions experiencing an outbreak instead of imposing nationwide or statewide lockdowns. We refer to these more focussed lockdowns as \textit{local lockdowns}.

For local lockdowns to be effective, it is crucial to enforce them at an early stage while the infections are still contained within the region of interest. This requires not only continuous monitoring of the daily case counts but also accurate forecasts to help local administration take appropriate action in advance. CoviHawkes uses a deep learning model to make such predictions at the national, state, and district level. The proposed tool learns from: \textbf{(i)} historical patterns in the case counts, \textbf{(ii)} information about the average movement of people in the region (hereafter referred to as mobility features), and \textbf{(iii)} the region's demographic information. The data used by our tool is publicly available and aggregated at an appropriate level to ensure that it cannot be traced back to individuals, thus preserving their privacy.

At the heart of CoviHawkes is a well-known stochastic process known as the \textit{Hawkes process}~\citep{Hawkes:1971:Spectra_of_Some_Self-Exciting_and_Mutually_Exciting_Point_Processes}. Loosely speaking, a Hawkes process characterizes the rate at which such events happen. (Here, A person getting infected with the virus can be thought of as an event). It defines a probability distribution over the occurrences of these events in a way that makes future events more likely if the rate has been high in the recent past. In other words, it models a scenario where future infections are more likely if many people have caught the disease in the recent past. For this reason, the Hawkes process is known as a \textit{self-exciting process}. Due to the nature of the spread of infectious diseases, a Hawkes process is a leading candidate for modeling such a phenomenon \citep{Chiang:EtAL:2021:Hawkes_process_modeling_of_COVID-19_with_mobility_leading_indicators_and_spatial_covariates, Garetto:EtAL:2021:A_time-modulated_Hawkes_process_to_model_the_spread_of_COVID-19_and_the_impact_of_countermeasures}. Our tool \textsf{CoviHawkes} uses deep neural networks to model the parameters of the Hawkes process as a function of the input features mentioned in the previous paragraph.

The remainder of this article is organized as follows: Section \ref{section:model} describes the underlying mathematical framework and our deep learning model in detail. Section \ref{section:short_term_predictions} establishes the validity of short-term forecasts generated by our tool using standard machine learning procedures for model validation. Section \ref{section:long_term_predictions} describes our approach for generating long-term forecasts and presents the corresponding results under various lockdown conditions. We wish to emphasize that \textsf{CoviHawkes} can notably predict short-term forecasts at the level of each district that makes it especially useful for policymakers in devising strategies for local lockdowns. We hope that such an approach would minimize the negative consequences of blanket lockdowns while still effectively containing the spread of the disease.


\section{CoviHawkes Model}
\label{section:model} 
In this section, we describe our method for a region of interest $\mathcal{R}$ that can be a district, state, or nation. 
Let $C(t) \in \mathbb{N}$ denote the number of new Covid-19 cases in the region $\mathcal{R}$ at time $t$. We use $\mathbf{m}(t) \in \mathbb{R}^{d_m}$ to denote a vector of mobility features for region $\mathcal{R}$ that describes the percentage change in various activities such as ``going to work'', ``staying at home'', ``grocery shopping'', etc, in this region on day $t$. Additionally, we will use $\mathcal{H}(t)$ to refer to the history of observed data till time $t$, i.e., $\mathcal{H}(t) = \lbrace (C(s), \mathbf{m}(s)) \rbrace_{s \leq t}$.

Conditioned on the history $\mathcal{H}(t - 1)$, our model assumes that $C(t)$ is a Poisson distributed random variable with mean $\lambda(t)$. Therefore, for all $c = 0, 1, 2, \dots,$
\begin{equation*}
    \mathrm{P}\left( C(t) = c \;\vert\; \mathcal{H}(t - 1) \right) = \frac{ \lambda (t)^{c} \exp(-\lambda(t))}{c\,!}.
\end{equation*}
Following the discrete-time Hawkes process literature, we model $\lambda(t)$ as
\begin{equation*}
    \lambda(t) = \mu + \sum_{i=1}^L w(L-i)  R(t-i) C(t-i).
\end{equation*}
Here, $\mu \in [0, \infty)$ is the base count rate for the region of interest, $L$ is the length of the time window within which past case counts affect the present case count, $R(t)$ is the \textit{reproduction number} of the virus, and $w(1), \dots, w(L)$ are weights assigned to the previous $L$ days such that $w(i) \geq 0$ and $\sum_{i=1}^L w(i) = 1$.

In other words, infections occur on day $t$ with a base rate of $\mu$. However, a person may also catch the virus from another person infected $i$ days ago ($i \leq L$). The probability of such a infection is proportional to the weight $w(L - i)$ assigned to that day and the reproduction number $R(t - i)$ of the virus $i$ days ago. The parameters $\mu$, $w(1), \dots, w(L)$, and $R(t)$ are learned directly from the data by maximizing the log-likelihood of the observations. We model $R(t)$ as a function of the past mobility data using an LSTM as described in Section \ref{subsection: R value Estimation}.

In the real world, a region has a finite population and it is rare for people to get infected twice. Moreover, vaccinated people are also less likely to get infected, and hence can be removed from the set of susceptible population members for modeling purposes \citep{Rizoiu:EtAL:2018:SIR-Hawkes_Linking_Epidemic_Models_and_Hawkes_Processes_to_Model_Diffusions_in_Finite_Populations}. To consider these factors, we use a discounted value of $\lambda(t)$ as shown below:
\begin{equation*}
    \tilde{\lambda}(t) = \left( 1 -\frac{ N(t-1) + V(t-1)}{N} \right) \lambda(t).
\end{equation*}
Here,  $N(t - 1) = \sum_{s = 1}^{t - 1} C(s)$ is the total number of people that have already been infected before time $t$, $V(t - 1)$ is the number of vaccinated people, and $N$ is the total size of the population in the region. As $N(t-1)$ and $V(t-1)$ increase, the discounted rate $\tilde{\lambda}(t)$ decreases. All parameters including the parameters of the LSTM that computes $R(t)$ are learned via maximum-likelihood estimation.


\subsection{Estimating the value of $R(t)$}
\label{subsection: R value Estimation}

$R(t)$ measures the rate at which the virus reproduces itself. A higher value of $R(t)$ entails faster spread of the disease and vice versa. Owing to the emergence of variants of the virus, the value of $R(t)$ changes over time and must be estimated periodically using recent trends in the data. To do so, we use an LSTM \citep{HochreiterSchmidhuber:1997:LongShortTermMemory}. Let $\mathbf{x}(t) \in \mathbb{R}^{(L + 1)d_m}$ be a vector obtained by concatenating the mobility vectors $\mathbf{m}(t - i - \Delta)$ for $i = 1, \dots, L$, and the vector of case counts in the previous $L$ days $[C(t - L), \dots, C(t - 1)]$. The mobility features used at time $t$ at delayed by an additional gap $\Delta > 0$ to take the incubation period of the disease into account. We use $d_m = 6$, $L = 28$, and $\Delta = 14$ in our experiments. $R(t)$ is then computed as:
\begin{align*}
    \mathbf{h}(t) = \mathrm{LSTM}(\mathbf{h}(t - 1), \mathbf{x}(t) ) \nonumber \\ 
    R(t) = \ln(1 + \exp(\mathbf{w}^\intercal \mathbf{h}(t) + b)).
\end{align*}
Here, $\mathbf{h}(t) \in \mathbb{R}^d$ is the hidden state of the LSTM at time $t$, and $\mathbf{w} \in \mathbb{R}^d$ and $b \in \mathbb{R}$ are learnable parameters.


\begin{figure}
    \centering
    \includegraphics[width=\linewidth]{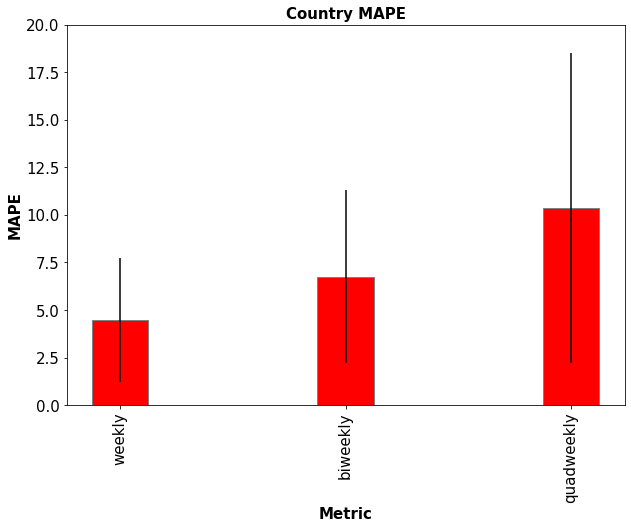}
    \caption{Country-Level MAPE scores for different sizes of forecasting window.}
    \label{fig:country-level forecast}
\end{figure}

\section{Short-term predictions}
\label{section:short_term_predictions}

For validating our short-term predictions, we use case counts $C(t)$ and mobility features $\mathbf{m}(t)$ collected between March 2, 2020, and July 20, 2021. We divide this data into two subsets: training data (March 2, 2020 to April 27, 2021) and validation data (April 27-July 20, 2021), and validate our model for forecasting windows of three sizes: $7$, $14$, and $28$ days. 

The validation period has $84$ days. To validate the model for a forecasting window of size $w$, we divide the validation period into $n$ possibly overlapping intervals of size $w$ each. Let $I^w_1, I^w_2, \dots, I^w_n$ be these intervals, then:
$$I^w_i = \lbrace t_{\mathrm{s}} + 7(i - 1) + k \rbrace_{k = 0}^{w-1},$$
where $t_s$ marks the start of the validation period. To make predictions for interval $I^w_i$, we independently train a model using all available data till the start of $I^w_i$ and compute our predictions for this interval. Let $\hat{C}^w_i(t)$ be the prediction made by such a model at time $t$. The error for interval $I^w_i$ is calculated as:
\begin{equation*}
    \psi(I^w_i) = \frac{\vert \sum_{t \in I^w_i} C(t) - \sum_{t \in I^w_i} \hat{C}^w_i(t) \vert}{\sum_{t \in I^w_i} C(t)} \times 100.
\end{equation*}
The error metric above is known as the Mean Absolute Percentage Error (MAPE). The process above is repeated for all $i = 1, \dots, n$, and the error of the model on a window of size $w$ is given by:
\begin{equation*}
    E(w) = \frac{1}{n} \sum_{i = 1}^n \psi(I^w_i).
\end{equation*}

\begin{figure*}
    \centering
    \subfloat[$w=7$]{\includegraphics[width=0.33\linewidth]{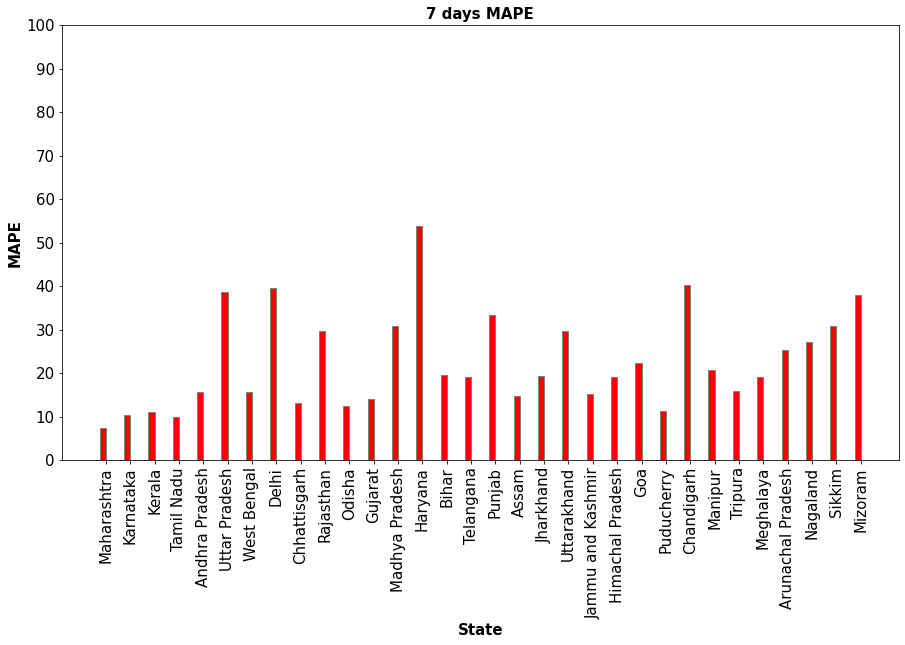}}%
    \subfloat[$w=14$]{\includegraphics[width=0.33\linewidth]{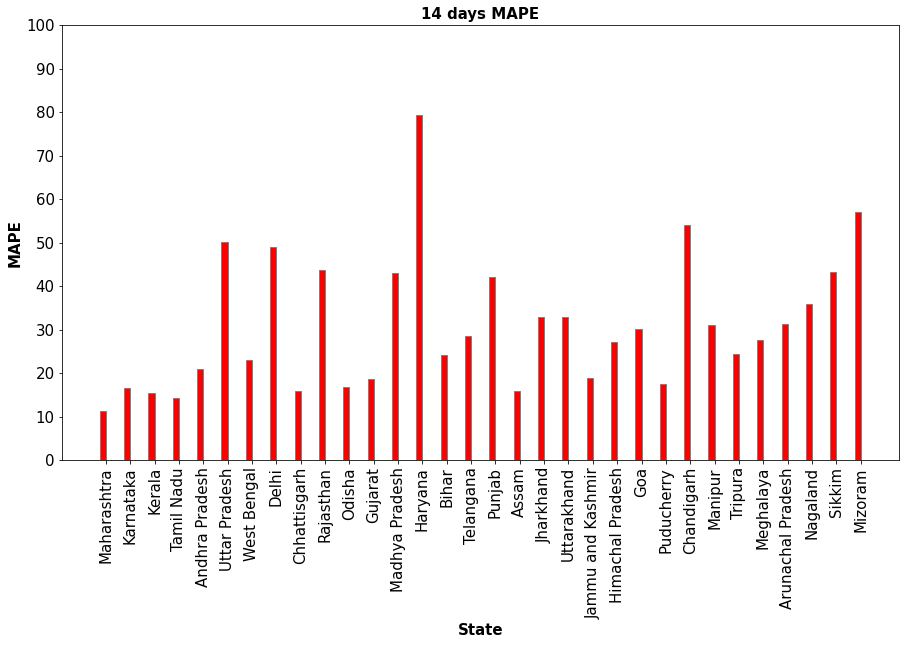}}%
    \subfloat[$w=28$]{\includegraphics[width=0.33\linewidth]{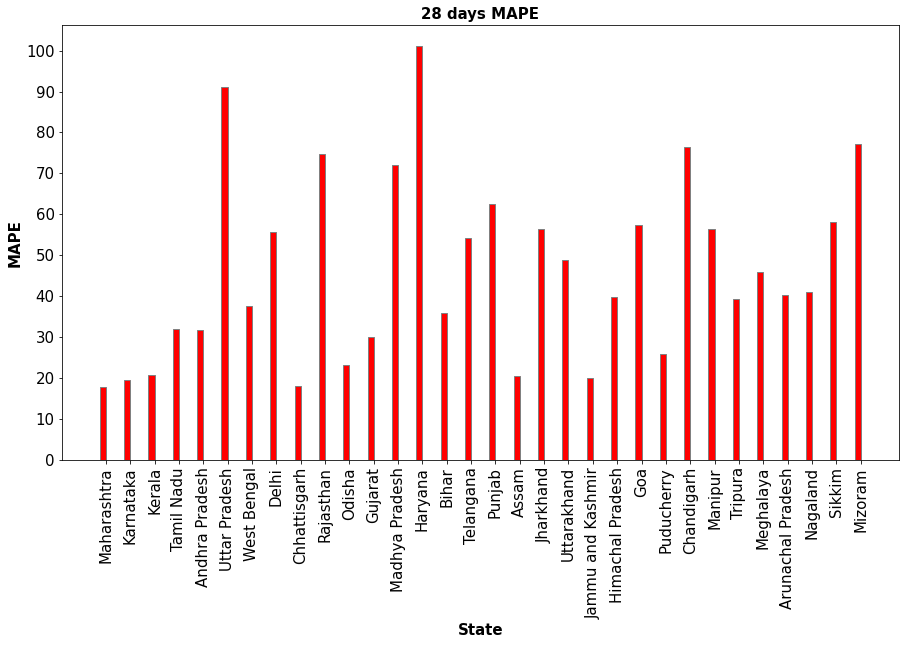}}
    \caption{MAPE for models trained at the state level for different states and values of forecast window size $w$.}
    \label{fig:state-level forecast}
\end{figure*}

\begin{figure*}
    \centering
    \subfloat[$w=7$]{\includegraphics[width=0.33\linewidth]{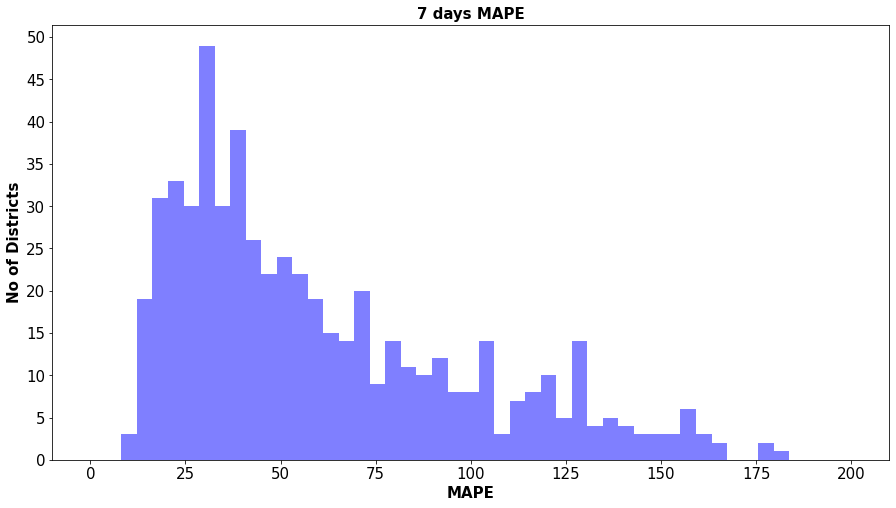}}%
    \subfloat[$w=14$]{\includegraphics[width=0.33\linewidth]{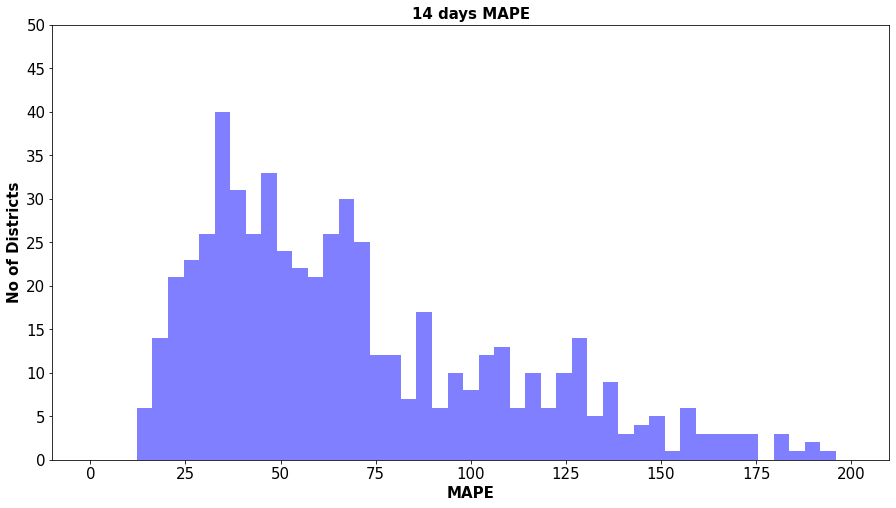}}%
    \subfloat[$w=28$]{\includegraphics[width=0.33\linewidth]{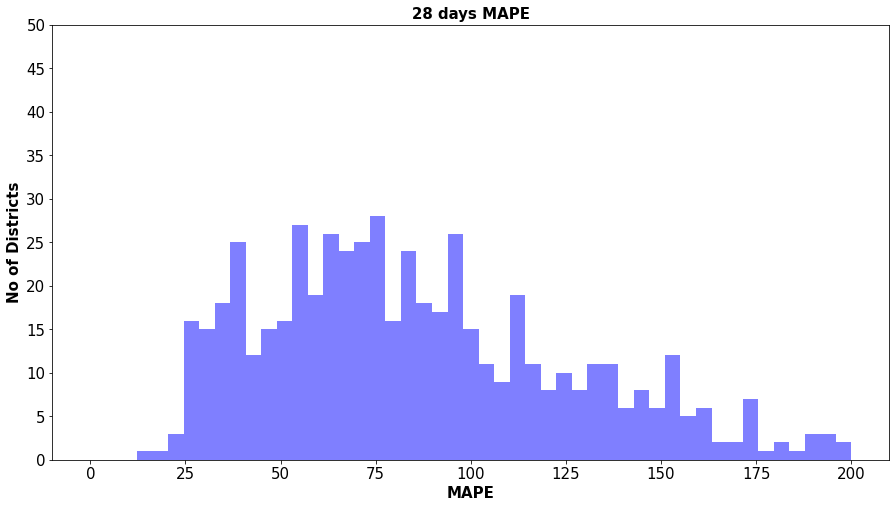}}
    \caption{Histogram of MAPE scores for models trained at district level for different values of forecast window size $w$.}
    \label{fig:district-level forecast}
\end{figure*}

Figure \ref{fig:country-level forecast} shows $E(w)$ for $w = 7$, $14$, and $28$ for the nation-level model that forecasts case counts for the country as a whole. Figure \ref{fig:state-level forecast} shows the same data for state-level models. As the number of districts is very large ($>500$), we use histograms in Figure \ref{fig:district-level forecast} to show the number of districts on the $y$-axis that have the corresponding MAPE scores on the $x$-axis. As before, Figure \ref{fig:district-level forecast} has histograms for different values of $w$. The latest forecasts based on this model are available on our website \url{https://sml.csa.iisc.ac.in/covihawkes}.


\begin{table*}
    \centering
    \begin{tabular}{>{\centering\arraybackslash}m{1.5cm} p{7cm} p{7cm}}
        \toprule
        \multicolumn{1}{>{\centering\arraybackslash}m{1.5cm}}{\textbf{Interval}} & \multicolumn{1}{>{\centering\arraybackslash}m{7cm}}{\textbf{Allowed}} & \multicolumn{1}{>{\centering\arraybackslash}m{7cm}}{\textbf{Not Allowed}} \\
        \midrule
        $I_s$ & \multicolumn{1}{p{7cm}}{Essential government services like defence, police, and power generation. Healthcare systems, related industries, and emergency services. Banks. Shops selling ration, meat, dairy, and animal fodder. Internet services. E-commerce. Media.} & \multicolumn{1}{p{7cm}}{All non-essential offices, shops, and industries. All transport (except for essential goods). All educational institutes, places of worship, and gatherings of any kind. Hospitality services.} \\ \hline
        $I_m$ & \multicolumn{1}{p{7cm}}{All businesses, some at limited capacity. Graded reopening of schools. Cinema/Theaters at 50\% capacity. Gatherings of up to 100 people in closed spaces with maximum occupancy of 50\%. Transportation.} & \multicolumn{1}{p{7cm}}{Large gatherings in indoor spaces. Different states have different restrictions.} \\ \hline
        $I_c$ & \multicolumn{2}{c}{Same as $I_m$, but with higher mobility values (see Figure \ref{fig:mobility_heatmap})} \\ \hline
        $I_n$ & \multicolumn{2}{c}{Everything open, business as usual} \\
        \bottomrule
    \end{tabular}
    \caption{Lockdown conditions. $I_s$: Strict lockdown, $I_m$: Unlock Phase 7, $I_c$: Current conditions, $I_n$: No lockdown. See Section \ref{section:long_term_predictions} for details about these conditions.}
    \label{table:lockdown_conditions}
\end{table*}

\section{Long-term predictions}
\label{section:long_term_predictions}

Recall that our model uses mobility features $\lbrace \mathbf{m}(t - \Delta - i) \rbrace_{i = 1}^L$ and case counts $\lbrace C(t - i) \rbrace_{i = 1}^L$ from the last few days to forecast the future value at time $t$. Thus, we can only generate a forecast at time $t$ if these feature values are available. To make predictions over a longer time horizon, we need appropriate proxies for the values of these features. The case counts can be boot-strapped, i.e., the model can treat its own past predictions as observed ground truth values. However, simulating the mobility values is much harder. Instead, we take a simplified approach as explained below.

We first identify four time intervals in the past that correspond to different lockdown conditions. The first interval, $I_s = [$March 25-April 14, 2020$]$, covers the time when there was a strict nationwide lockdown. The second interval, $I_m = [$December 13-19, 2020$]$, falls in the seventh \textit{unlock} phase in India where most, though not all, of the restrictions, were lifted. The third interval, $I_n = [$February 15-March 3, 2020$]$ corresponds to the time before the pandemic when there was no lockdown. Finally, $I_c = [$August 13-19, 2021$]$, corresponds to the current mobility conditions. Table \ref{table:lockdown_conditions} summarizes the conditions under these intervals.

We generate four long-term forecasts, one each for the mobility conditions from the intervals $I_s$, $I_m$, $I_n$, and $I_c$. To generate forecasts corresponding to $I_{\mathrm{x}}$ ($\mathrm{x} \in \lbrace s, m, n, c \rbrace$), we compute the average value of mobility features for each weekday $i = 1, \dots, 7$ in $I_{\mathrm{x}}$. Let $D(i)$ be the set of all $i^{th}$ weekdays between February 14, 2020 and August 20, 2021. For example, $D(1)$ will be the set of all Sundays in this interval. Then, define $\mathbf{m}_{\mathrm{x}}(i)$ as,
\begin{equation*}
    \mathbf{m}_{\mathrm{x}}(i) = \frac{1}{\vert I_{\mathrm{x}} \cap D(i)  \vert} \sum_{t \in I_{\mathrm{x}}} \mathbf{1}\lbrace t \in D(i) \rbrace \mathbf{m}(t).
\end{equation*}
Whenever we require the mobility value for a day $t$ in the future that has not been observed, we use $\mathbf{m}_{\mathrm{x}}(i)$ for an appropriate $i$ chosen based on the day of the week at time $t$. This, together with bootstrapped count values, enables us to use our model to generate long-term forecasts.

\begin{figure}[t]
    \centering
    \includegraphics[width=\linewidth]{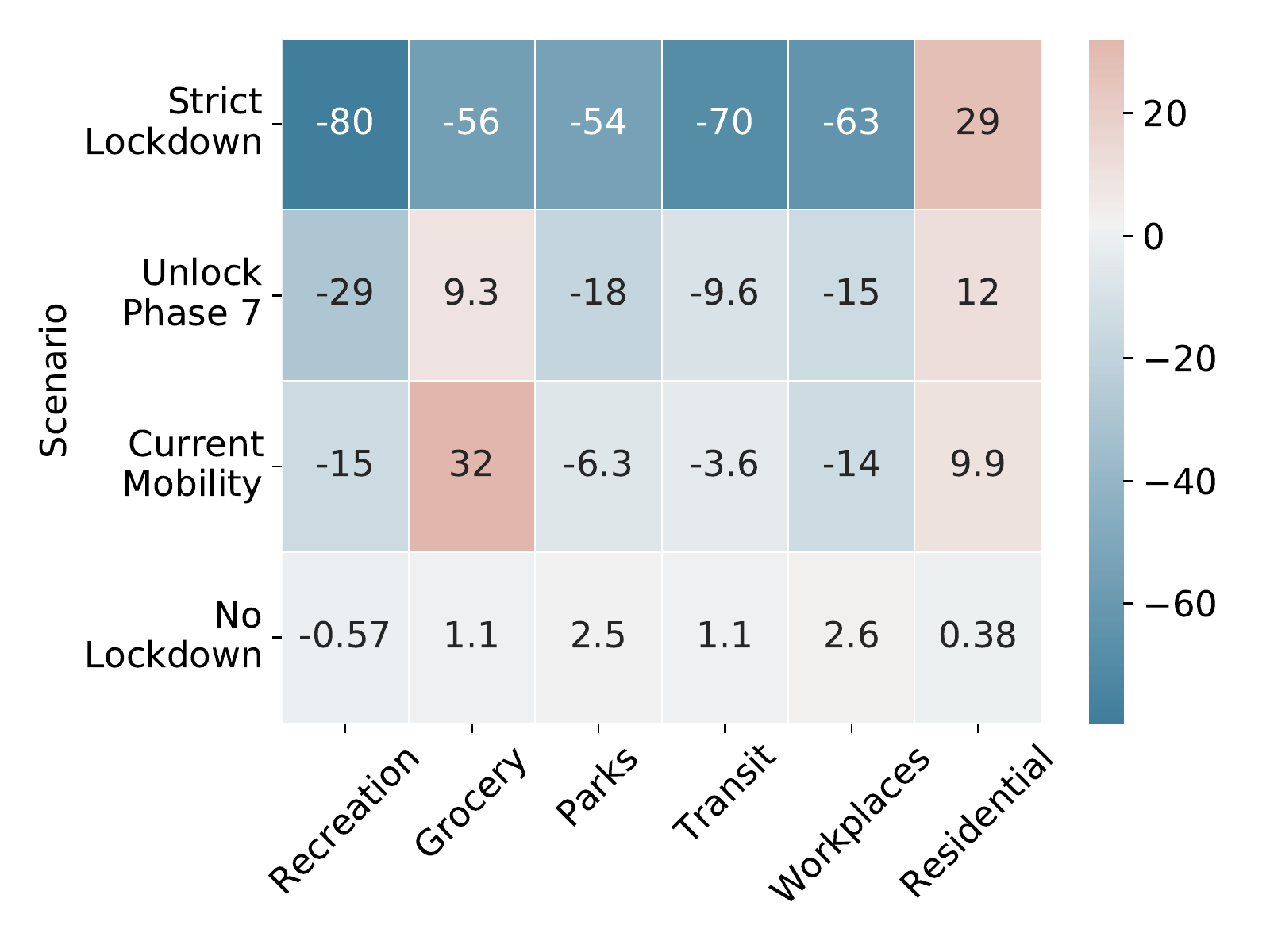}
    \caption{Heatmap showing average mobility values under various lockdown conditions. The columns correspond to the six mobility features that we use. Negative values indicate less mobility (the values represent percentage change as compared to a pre-pandemic baseline).}
    \label{fig:mobility_heatmap}
\end{figure}

\begin{figure}[t]
    \centering
    \includegraphics[width=\linewidth]{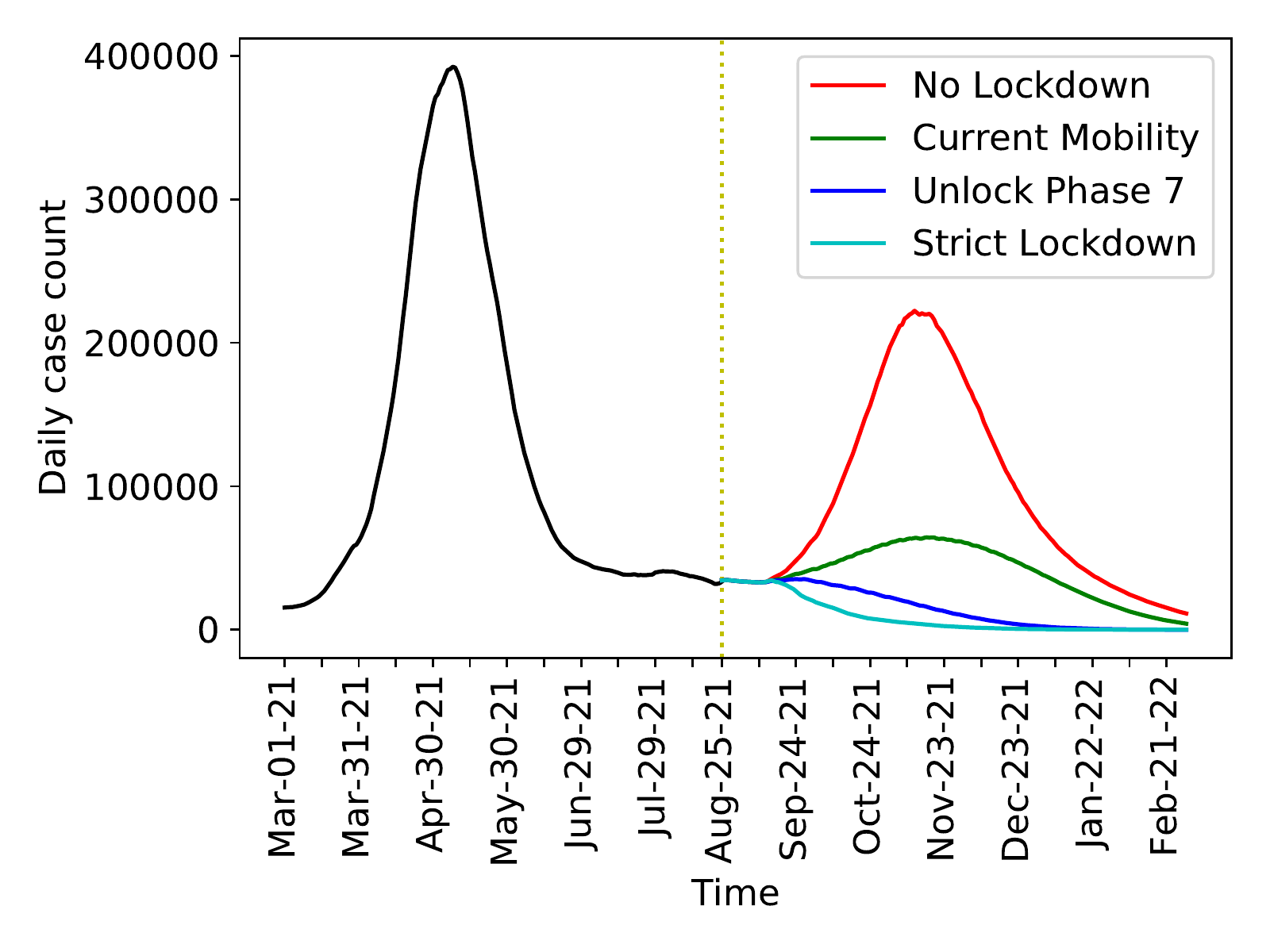}
    \caption{Nation-level long-term forecast.}
    \label{fig:country-level-longterm-forecast}
\end{figure}

Figure \ref{fig:mobility_heatmap} shows the average mobility values $\frac{1}{7} \sum_{i=1}^7 \mathbf{m}_{\mathrm{x}}(i)$ for various lockdown conditions and Figure \ref{fig:country-level-longterm-forecast} shows the long-term forecast at the nation-level under these conditions. One can see from Figure \ref{fig:mobility_heatmap} that the current conditions have a higher mobility value as compared to the seventh unlock phase, despite the restrictions being similar in these two phases (see Table \ref{table:lockdown_conditions}). Consequently, the model predicts a sharper rise in cases under $I_c$ as compared to $I_m$. One can also see that the model predicts a significant third wave if the mobility returns to the pre-pandemic conditions ($I_n$). Interestingly, the model also indicates that a mild lockdown (as in $I_m$) would be almost as good as a very strict lockdown (as in $I_s$).

As is the case with every statistical model, \textsf{CoviHawkes} also makes several assumptions. For example, we train models independently for states and districts and do not consider the movement of people across these regions due to the lack of this data. Similarly, the model does not consider breakthrough infections, reinfections, or emergence of new variants of the virus in the long term. While our short-term forecasts are rigorously validated, validating long-term forecasts can be challenging. In particular, long-term forecast uses proxy values for the features as described above, and hence the errors compound over time. Moreover, usage of average mobility features ignores events like festivals and/or other unforeseen mass gatherings that are likely to trigger an uptick in the infection rate. As such, one must keep these caveats in mind while using these long-term predictions.


\section{Conclusion}
In this article, we introduced an AI tool for forecasting Covid-19 case counts in India at nation, state, and district levels. This model is based on Hawkes process, which is suitable for modeling the spread of infectious diseases due to its self-exciting nature. We rigorously validated the short-term predictions made by our model using standard validation procedures for time-series data. We also used our model to generate long-term forecast at the national level to provide an indication of case counts under different lockdown conditions. We hope that this model, especially its short-term forecasts at the district level, will be useful for policymakers in developing strategies for local lockdowns.


\subsection*{Acknowledgements}
Many members of StatsML group at IISc contributed to this work. We especially acknowledge contributions of Parag, Abhishek, Zaid, Tejas and Rahul. 

\bibliographystyle{apalike}
\bibliography{biblio}

\end{document}